\documentclass{article}

\usepackage{graphicx} 
\usepackage{subfigure} 

\usepackage{natbib}

\usepackage{algorithm}
\usepackage{algorithmic}

\usepackage{hyperref}

\usepackage[accepted]{icml2013}

\icmltitlerunning{The Orchive : Data mining a massive bioacoustic archive}

\begin{document} 

\twocolumn[
\icmltitle{The Orchive : Data mining a massive bioacoustic archive}

\icmlauthor{Steven Ness}{sness@uvic.ca}
\icmladdress{Department of Computer Science,
            University of Victoria, Canada}
\icmlauthor{Helena Symonds}{info@orcalab.org}
\icmladdress{OrcaLab, P.O. Box 510
            Alert Bay, BC, Canada}
\icmlauthor{Paul Spong}{info@orcalab.org}
\icmladdress{OrcaLab, P.O. Box 510
            Alert Bay, BC, Canada}
\icmlauthor{George Tzanetakis}{gtzan@cs.uvic.ca}
\icmladdress{Department of Computer Science,
            University of Victoria, Canada}

\icmlkeywords{bioacoustics, machine learning, whale, orca}

\vskip 0.3in
]

\begin{abstract} 
  The Orchive is a large collection of over 20,000 hours of audio
  recordings from the OrcaLab research facility located off the
  northern tip of Vancouver Island.  It contains recorded orca
  vocalizations from the 1980 to the present time and is one of the
  largest resources of bioacoustic data in the world.  We have
  developed a web-based interface that allows researchers to listen to
  these recordings, view waveform and spectral representations of the
  audio, label clips with annotations, and view the results of machine
  learning classifiers based on automatic audio features extraction.
  In this paper we describe such classifiers that discriminate between
  background noise, orca calls, and the voice notes that are present
  in most of the tapes.  Furthermore we show classification results
  for individual calls based on a previously existing orca call
  catalog. We have also experimentally investigated the scalability of
  classifiers over the entire Orchive.

\end{abstract}

\section{Introduction}
\label{introduction}
The Orchive is a large archive containing over 20,000 hours of
recordings from the Orcalab research station. These recordings were
made using a network of hydrophones and originally stored on analog
cassette tapes. OrcaLab is a research station on Hanson Island which
is located at the north part of Vancouver Island on the west coast of
Canada. It has been in continuous operation since 1980.  It was
designed as a land based station in order to reduce the impact on the
orcas under study, as the noise and disturbance from boats affects the
orcas in observable but currently unquantified ways.  In collaboration
with OrcaLab, we have digitized the tapes and have made these
recordings available to the scientific community through the Orchive
website (\hyperref[http://orchive.cs.uvic.ca]{http://orchive.cs.uvic.ca}).

\begin{figure}[t]
\centering
\includegraphics[width=\columnwidth]{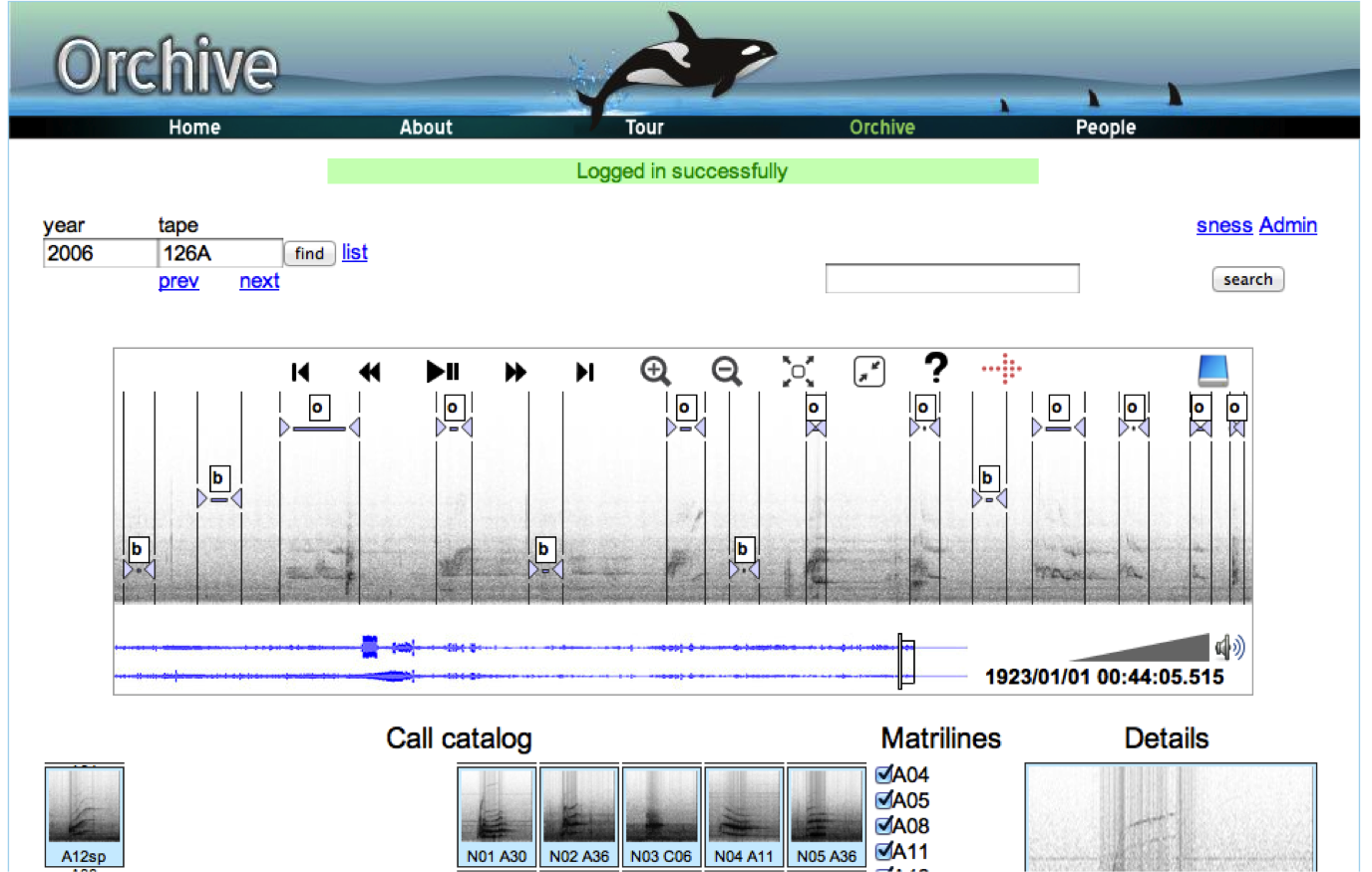}
\caption{Annotated audio from from the Orchive}
\label{fig:dm_orchive}
\end{figure}

Over the past 5 years, a number of orca researchers using our website
have added over 18,000 clip annotations to our database.  A small
section of annotated audio from the Orchive is shown in Figure
\ref{fig:dm_orchive}.  These clip annotation are of two main types:
The first is clips that differentiate background noise from orca calls
and from the voice notes of the researchers that collected the data.
The second type of clip annotations classify orca vocalizations into
different calls.  Orcas make three types of vocalizations,
echolocation clicks, whistles and pulsed calls.  The pulsed calls are
highly conserved stereotyped vocalizations which have been classified
into a catalog of over 52 different calls by John Ford \cite{ford87}.
Of the 18,000 annotations currently in the Orchive, 3000 are of these
individually classified calls.  In addition, we have a curated call
catalog containing 384 different recordings of different calls
vocalized by a variety of different pods and matrilines. This catalog
is used for training the annotators. 

Many parts of the recordings contain boat noise which makes
identifying orca calls both difficult and tiring. In addition, the
size of the Orchive makes full human annotation practically
impossible. Therefore we have explored machine learning
approaches to the task. One data mining task is to segment and label the
recordings with the labels background, orca, voice. Another is to
subsequently classify the orca calls into the classes specified in the
call catalog.

\section{Related Work}
\label{relatedWork}

Audio feature extraction is the first step in classifying audio using
machine learning algorithms.  Mel-Frequency Cepstral Coefficients
\cite{Logan00melfrequency} (MFCC) have been widely used for this
purpose.  MFCCs have also been used in bioacoustics, and have been
used to classify bird songs \cite{changhsing07} and orca calls
\cite{ness08}.  In this work we also use MFCCs, but supplement them
with other audio features including Centroid frequency, Rolloff
frequency, Flux, and Zero Crossings.

Our system uses two types of web based interfaces.  The first are
tools aimed at expert users, and the second are simpler interfaces
designed for crowdsourcing the annotation. There are a number of tools
that experts use to segment and analyze audio and specifically
bioacoustic data.  One of the most popular is Raven
\mbox{(\hyperref[http://www.birds.cornell.edu/raven]{http://www.birds.cornell.edu/raven})},
a toolkit developed at the Cornell Lab of Ornithology.  The biggest
difference our system compared to systems such as Raven is that our
system web-based, can more easily view and analyze large amounts of
data.

\section{System Overview}
\label{systemDescription}

\begin{figure}
\centering
\includegraphics[width=80mm]{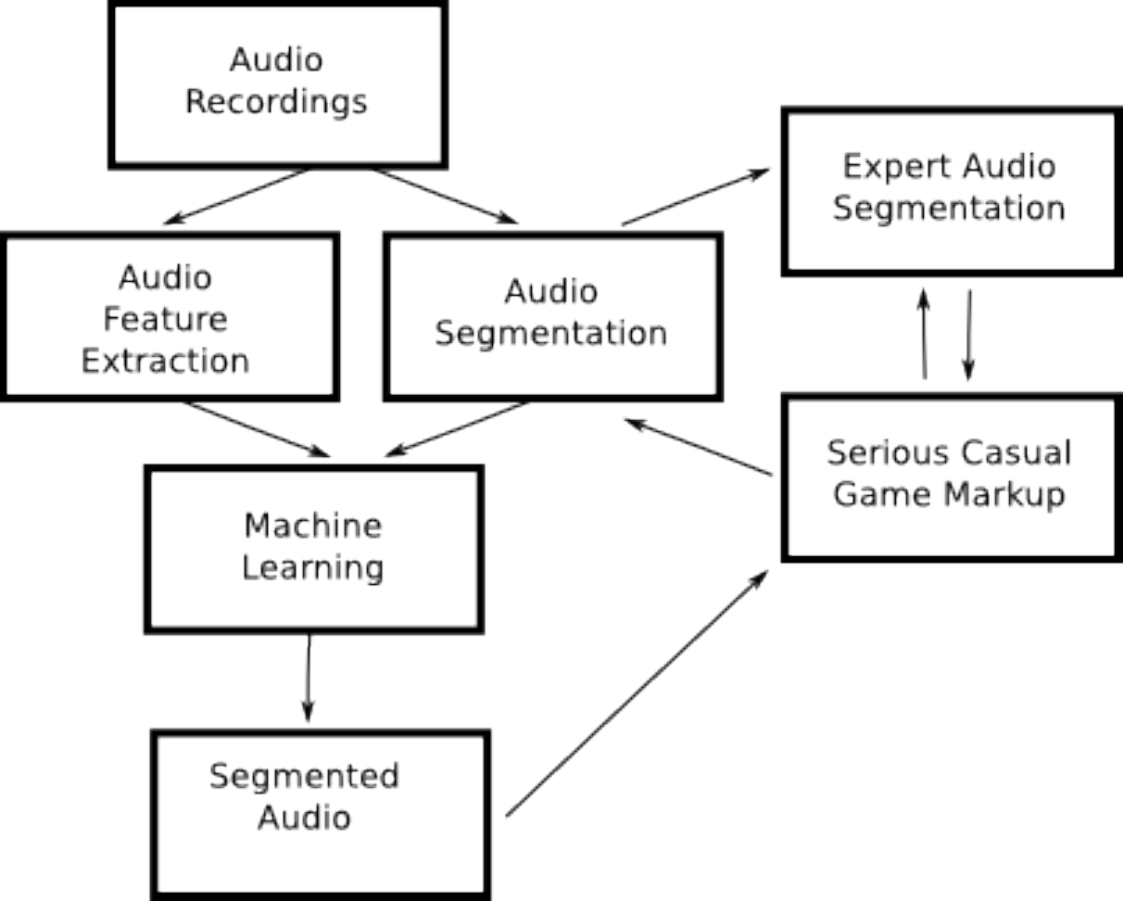}
\caption{System Diagram}

\label{fig:systemDiagram} 
\end{figure}

We have developed a collaborative web interface that allows expert
researchers to listen to, view and annotate large collections of audio
data.  The system also supports a variety of audio feature extraction
and machine learning algorithms, and enables users to view the results
of these algorithms.  
A diagram of this system is shown in Figure \ref{fig:systemDiagram}. 

For audio features we use the Marsyas\cite{marsyas} Music Information
Retrieval system.  Marsyas allows us to perform both audio feature
extraction and machine learning on audio data directly.  


In order to efficiently analyze large audio archives we utilize
distributed computing. There are many systems for distributing
computation. We currently use the Portable Batch System (PBS)
\cite{henderson95}, a grid-computing system where similar data can be
processed in parallel by a large number of computers.


\section{Experimental Results}
\label{experimentalResults}

\subsection{Audio Feature Extraction Parameters}

The first set of parameters that needed to be optimized were the
Window Size and Hop Size of the Digital Signal Processing (DSP)
algorithms that take the input audio and calculate spectral
information from them, the fundamental basis for which is the Fast
Fourier Transform (FFT) algorithm.  The length of time over which to
calculate the statistical properties of the features, this is known in
bextract as the ``memory'' and corresponds to the number of frames of
features that are accumulated.  We ran this on a 600 second audio
dataset labeled as orca, background and voice with equal lengths of
each label.  In this dataset, the voice was trimmed by hand, the orca
consisted of the middle 0.023 seconds of approximately 10,000 clips,
and the background consisted of 0.15 seconds of approximately 1300
clips.  The results for this are shown in Table \ref{table:dspParams}.
From this we can see that as we go to longer window sizes, the
classification performance increases, and as we go to longer
accumulation window sizes, the performance also increases. For the
remaining experiments we use these optimal settings.

\begin{table}
\begin{tabular}{|r|r|r|r|r|r|}
\hline
 winsize  &  hopsize  &  memory  &  $\#$ correct  \\
\hline
 20 &  512 &  256 &     70.16 \\
 20 & 1024 &  512 &     71.88 \\
 20 & 2048 & 1024 &     74.17 \\
 20 & 4096 & 2048 &     73.38 \\
\hline
 40 &  512 &  256 &     72.94 \\
 40 & 1024 &  512 &     75.67 \\
 40 & 2048 & 1024 &     78.29 \\
 40 & 4096 & 2048 &     80.58 \\
\hline
 80 &  512 &  256 &     76.53 \\
 80 & 1024 &  512 &     78.39 \\
 80 & 2048 & 1024 &     81.88 \\
 80 & 4096 & 2048 &     85.72 \\
\hline
\end{tabular}
\caption{In this table results of a systematic parameter search
  through different DSP parameters is shown.  winsize is the window
  size of the FFT in samples, and hop size is the number of samples
  skipped between each successive application of the FFT.  memsize
  refers to the number of FFT frames on which the mean and standard
  deviation are determined.}
\label{table:dspParams}
\end{table}

\subsection{Orca/Background/Voice Classification}

The first task we investigate is the classification of audio into
three classes: orca, background, and human voice.  In order to test
the different distributed audio classification systems we first
generated a set of training and testing data, one of these was a set
of calls from the curated call catalog with silence removed, and the
other was an entire 45 minute recording from the Orchive which had
been annotated by an orca researcher. In a previous paper
\cite{ness08}, we were able to obtain a classification performance of
82\% when using a SVM classifier on hand labeled data.  We looked in
more detail at the training data, and found that there was a small
amount of silence before and after the vocalization.  The results can
be found in the first line of table \ref{table:handTrimmed} and had
93.5\% of the instances classified correctly.  This large jump in
performance was unexpected but easily understood, because if feature
vectors of silence are labeled as orca, this will cause issues for the
classifier.  We then took a 4 minute region of orca calls and voice
notes and removed all the silences from both of them, for this we
obtained a classification accuracy of 96.1\% when looking at the call
catalog dataset, and 95.0\% when looking at the annotated recording.

However, this process of hand trimming recordings would be unfeasible
to do on the entire 18,000 current annotations.  For this, we instead
tested a procedure where we extracted a small section of audio from
the middle of each clip where it was most probable that the orca call
would be found.  

We then extracted audio features from these sections of audio using
Marsyas.  Marsyas has a wide variety of audio features that it can
calculate, including MFCCs, number of zero crossings per window and
various high level descriptions of the spectrum including the centroid
(center of mass of the spectrum), rolloff (the frequency for which the
sum of magnitudes of its lower frequencies are equal to percentage of
the sum of magnitudes of its higher frequencies) and the flux (the
norm of the difference vector between two successive magnitue/power
spectra).  We tried different combinations of these, and found that
using all of these features gave the best performance.  All subsequent
results in this paper use all of these features.

To classify these features, we used a Sequential Minimal Optimization
implementation of a Support Vector Machine classifier \cite{platt98},
an algorithm which had shown its effectiveness in our previous work
\cite{ness08} in this problem domain.

The results for this procedure for a clip of 0.023
seconds from the middle of each orca call was 96.5\% and for the
recording from the Orchive, the accuracy was 93.4\%.

\begin{table}
\begin{tabular}{|l|c|l|l|r|r|}
\hline
Training       & length  & \% corr.   & \% corr.  & \% corr.  \\
dataset        &  (sec)  &  10-fold     &   (calls)   &   (442A)    \\
\hline
hand-10sec     &    30   &   99.4       &   93.5      &   93.1     \\
hand-4min      &    720  &   99.9       &   96.1      &   95.0   \\
ms 100         &    300  &   99.9       &   96.5      &   93.4   \\
\hline
\end{tabular}
\caption{Classification results with hand trimmed orca vocalizations
  using bextract using an SMO SVM classifier.}
\label{table:handTrimmed}
\end{table}

\subsection{Call classification}

Using the Orchive interface we created a collection of 197 calls of 6
classes, these included the common calls ``N1'', ``N3'', ``N4'',
``N7'', ``N9'' and ``N47''.  Audio features for each 20ms audio frame
of these files were generated, these included the MFCC coefficients,
Centroid, Rolloff, Flux and Zero crossings as described and justified
in the previous section.  The mean and standard deviation for each of
these features were then calculated and were output as a .arff file.
The SMO SVM classifier produced gave an accuracy of 98.5\% accuracy on
this set of calls, and the confusion matrix for this is shown in Table
\ref{table:orchiveConfusionMatrix}.




\begin{table}
\centering
\begin{tabular}{|c|cccc|} 
\hline
   &   N1   &   N4   &   N7   &   N9   \\
\hline
N1 & 1726  &   0     &   0    &   0    \\
N4 &   12  & 2858    &   0    &   0    \\
N7 &    0  &   2     & 1297   &   59   \\
N9 &    0  &   0     &  70    & 3231   \\
\hline
\end{tabular}
\caption{Confusion matrix for 10-fold crossvalidation with SVM
  classifier on labelled calls from Orchive.}
\label{table:orchiveConfusionMatrix}
\end{table}

\subsection{Performance}

In order to investigate the performance of the classification of
recordings into Orca, Background and Voice, we trained a SVM with a
section of 30 and 240 seconds of hand trimmed data using the bextract
program in Marsyas.  We then used the sfplugin program in Marsyas to
classify all the recordings in the Orchive on the Hermes/Nestor
cluster, part of the Westgrid computational resource.  For this we
divided the data into sets of 1\%, 5\%, 10\% and 100\% of the Orchive.
The timing results of these datasets run on 10 computers are shown in
Table \ref{table:performance}.  From this we can see that the
classifier that had more data took longer to classify, and that the
speedup from taking samples of the data was almost linear.

\begin{table}
\centering
\begin{tabular}{|c|c|c|} 
\hline
Training data & \% of Orchive & Run time \\
 (sec)        &               & (d:h:m:s) \\
\hline
30            &      1      &    00:00:05:18      	 \\
30            &      5      &    00:00:25:20     \\
30            &      10     &    00:00:50:58    \\
30            &      100    &    00:09:01:05       \\
\hline
240           &      1      &    00:06:16      \\
240           &      5      &    00:00:31:21       \\
240           &      10     &    00:04:47:12     \\
240           &      100    &    02:04:18:32    \\
\hline
\end{tabular}
\caption{Performance results of timing on subsets of the entire
  Orchive dataset.}
\label{table:performance}
\end{table}

\section{Conclusion}
\label{conclusion}

In this paper we described a system that allows orca researchers to
listen to, view and annotate the large amount of audio data in the
Orchive.  The system also allows researchers to run and view the
results of audio feature extraction and machine learning algorithms on
this data.

We investigated the performance of different parameters for the audio
feature extraction process and showed that in general, large window
sizes were beneficial, and that increasing the length of time that
statistics were taken over the data was also beneficial.  We showed
that by carefully hand editing clips to remove silence was very
useful, and boosted performance from around 90\% to 96\% on actual
recordings.  We then used these classifiers on a cluster to classify
all the recordings in the Orchive into the classes, Orca, Background
and Voice.  The performance of call classification was also good, with
a classification accuracy of 98.5\% using a collection of 197 calls
culled from the Orchive.  The calls most often misclassified were the
N7 and N9 calls, and these are also difficult for non-experts in orca
vocalizations to differentiate.

\bibliography{icml2013}
\bibliographystyle{icml2013}

\end{document}